\documentclass[conference]{IEEEtran}
\IEEEoverridecommandlockouts
\usepackage{cite}
\usepackage{amsmath,amssymb,amsfonts}
\usepackage{algorithmic}
\usepackage{graphicx}
\usepackage{textcomp}
\usepackage{xcolor}
\usepackage{newtxmath}
\usepackage{algorithmic}
\usepackage{algorithm}
\usepackage{multirow}
\usepackage{hyperref}

\def\BibTeX{{\rm B\kern-.05em{\sc i\kern-.025em b}\kern-.08em
    T\kern-.1667em\lower.7ex\hbox{E}\kern-.125emX}}

\makeatletter
\newcommand{\linebreakand}{%
  \end{@IEEEauthorhalign}
  \hfill\mbox{}\par
  \mbox{}\hfill\begin{@IEEEauthorhalign}
}
\makeatother

\begin{document}

\title{
Temporal Gaussian Copula For Clinical Multivariate Time Series Data Imputation \\
\thanks{*Lin Chen and Yuwen Chen are the corresponding authors. This research is supported in part by the National Nature Science Foundation of China (82030066 and 62371438) and also by the Chongqing Municipal Education Commission (HZ2021008, 24XJC630014, HZ2021017)}
}

\author{
\IEEEauthorblockN{Ye Su}
\textit{Chongqing Institute of }\\
\textit{Green and Intelligent Technology, CAS} \\
\textit{University of Chinese Academy of Sciences}\\
Chongqing, China\\
suye241@mails.ucas.ac.cn
\and
\IEEEauthorblockN{Hezhe Qiao}
\textit{Singapore Management University} \\
Singapore, Singapore \\
hezheqiao.2022@phdcs.smu.edu.sg
\and
\IEEEauthorblockN{ Di Wu}
\textit{Southwest University} \\
Chongqing, China\\
wudi.cigit@gmail.com
\linebreakand 
\IEEEauthorblockN{Yuwen Chen$^{\ast}$}
\textit{Chongqing Institute of }\\
\textit{Green and Intelligent Technology, CAS} \\
Chongqing, China\\
chenyuwen@cigit.ac.cn
\and
\IEEEauthorblockN{Lin Chen$^{\ast}$}
\textit{Chongqing Institute of }\\
\textit{Green and Intelligent Technology, CAS} \\
Chongqing, China\\
chenlin@cigit.ac.cn
}

\maketitle

 \begin{abstract}

 The imputation of the Multivariate time series (MTS) is particularly challenging since the MTS typically contains irregular patterns of missing values due to various factors such as instrument failures, interference from irrelevant data, and privacy regulations. Existing statistical methods and deep learning methods have shown promising results in time series imputation. In this paper, we propose a Temporal Gaussian Copula Model (TGC) for three-order MTS imputation. The key idea is to leverage the Gaussian Copula to explore the cross-variable and temporal relationships based on the latent Gaussian representation. 
Subsequently, we employ an Expectation-Maximization (EM) algorithm to improve robustness in managing data with varying missing rates.
Comprehensive experiments were conducted on three real-world MTS datasets. The results demonstrate that our TGC substantially outperforms the state-of-the-art imputation methods. Additionally, the TGC model exhibits stronger robustness to the varying missing ratios in the test dataset. Our code is available at https://github.com/MVL-Lab/TGC-MTS.

\end{abstract}


\section{Introduction}
\begin{figure*}[htbp]
	\centerline{\includegraphics[width=1.0\linewidth]{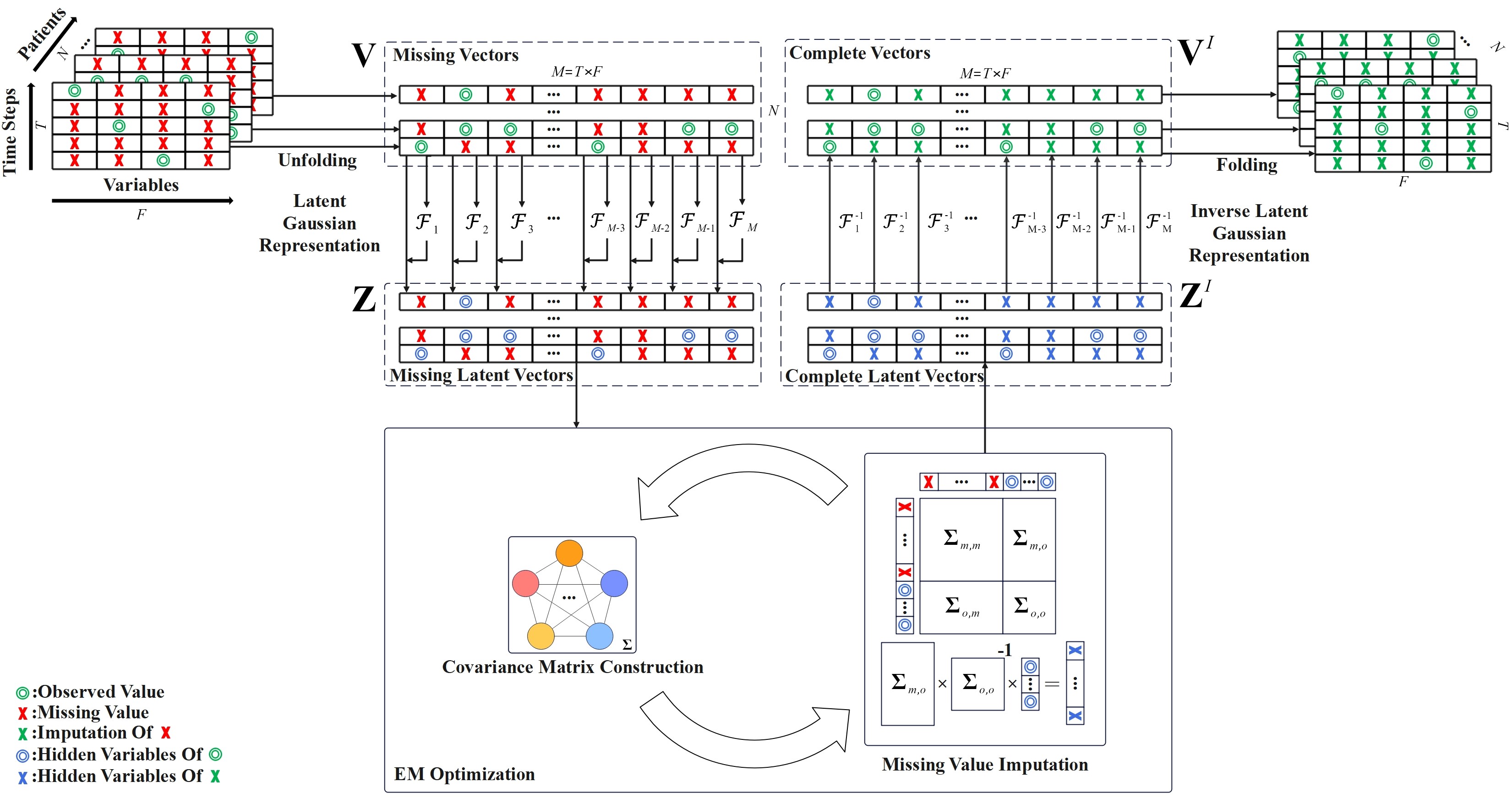}}
	\caption{The framework of TGC.}
	\label{framework}
\end{figure*}
As a prevalent data in Electronic Health Records (EHR), clinical multivariate time series (MTS) data is widely used for various tasks for biomedical analysis, such as evaluating treatment effects \cite{liu2023estimating},  morbidity forecasts \cite{nowroozilarki2021real}, and mortality projections \cite{hou2020predicting, bergquist2024evaluation}.  However, clinical MTS in the real world often contains many missing values due to various factors such as instrument failures, interference from irrelevant data, and privacy regulations \cite{ren2023review}, which presents a significant challenge to the application of existing machine learning methods and further impacts the accuracy of downstream analyses and decision-making.  Therefore, developing an accurate and effective strategy to handle missing data in MTS is essential for accurate clinical diagnosis.

While existing statistical and deep learning approaches \cite{liu2022scinet, du2023saits, liu2024koopa} have shown promise in clinical MTS imputation, several challenges remain: (1) The complex relationships among clinical values at different time points are not fully captured. (2) Irregular and non-continuous recording intervals among patients can bias EHR data distribution, and current methods often struggle with varying sampling densities, resulting in suboptimal performance.

To tackle these limitations, we developed a cross-variable and longitudinal imputation approach, named Temporal Gaussian Copula (TGC), based on a Gaussian Copula operation to fully explore the complex relation between different clinical values and time-points. We utilize a conditional expectation formula based on EM optimization to ensure that all observed values are fully leveraged in imputing missing values, thereby accommodating any sampling density. Specifically, we first unfold the 3D clinical data as a single matrix by concatenating the clinical longitudinal information of the patients. Then, we employ the Gaussian copula to estimate complex multivariate distributions of variable-by-time correlations through the feature transformations of the latent Gaussian vectors. Finally, the model was solved using an effective approximate Expectation-Maximization (EM) algorithm, and the missing value was imputed using a defined inverse potential Gaussian vector transform. 
We conducted extensive experiments on three real-world medical MTS datasets at varying missing rates to evaluate the performance of TGC. The experimental results demonstrate that the TGC model exhibits the best average performance across different missing rates.

\section{Methodology}
The overview of TGC is illustrated in Fig \ref{framework}. The TGC is primarily divided into four modules: (1) Three-order Tensor Unfolding. (2) Latent Gaussian Representation. (3) Covariance Matrix Construction. (4) Missing Value Estimation and Imputation.  Subsequently, a detailed exposition of each constituent of the TGC will be presented.

\subsection{Three-order Tensor Unfolding} 
We assume the MTS for a specific patient ${\bf{X}}_i \in \mathbb{R}^{T \times F}$, following the matrix Gaussian distribution with the parameters including mean matrix $\mathbf{M} \in \mathbb{R}^{T \times F}$,  temporal covariance $\mathbf{U} \in \mathbb{R}^{T \times T}$, and  covariance matrix $\mathbf{C} \in \mathbb{R}^{F \times F}$, which is formulated as the following:
 \begin{align}
     &\mathcal{P}({{\bf{X}}_i}\mid {\bf{M}},{\bf{U}},{\bf{C}})\nonumber\\
&=(2 \pi)^{-\frac{TF}{2}}|{\bf{C}}|^{-\frac{T}{2}}|{{\bf{U}}}|^{-\frac{F}{2}}e^{-\frac{1}{2} \operatorname{Tr}\left[{{\bf{C}}}^{-1}({\bf{X}}_i-{{\bf{M}}})^{\top} {{\bf{U}}}^{-1}({\bf{X}}_i-{{\bf{M}}})\right]}.\label{MN}
 \end{align}

Thus, we can establish the temporal cross-variable correlations for missing data estimation by computing the covariance matrix $\mathbf{\Sigma}$, using the matrices $\mathbf{C}$ and $\mathbf{U}$.

However, estimating high-dimensional parameters of $\mathbf{M}$, $\mathbf{C}$, and $\mathbf{U}$ is challenging due to the complexities inherent in matrix Gaussian. Therefore, to apply  a simple yet effective missing data estimation approach, we can unfold the 3D tensor ${\cal{X}}$ into a single 2D matrix $\mathbf{V}\in \mathbb{R}^{N\times M}$ which is formulated as: 
\begin{align}
    \mathbf{V}=\{vec({\mathbf{X}}_1),vec({\mathbf{X}}_2),...,vec({\mathbf{X}}_N)\},\label{expand}
\end{align}
where $M=T \times F$, and $vec(\mathbf{X}_i)$ is the vectorization operator that transforms matrix $\mathbf{X}_i$ into a column vector. Given each ${\mathbf{v}}_i \in {\mathbb{R}}^{M}$ follows the multivariate normal distribution, which is formulated as the following: 
\begin{align}
   \mathcal{P}({\bf{v}}_i|\mathbf{\mu},\mathbf{\Sigma})=(2\pi)^{-\frac{M}{2}}\lvert{\mathbf{\Sigma}}\rvert^{-\frac{1}{2}}e^{-\frac{1}{2} ({\bf{v}}_i-\mathbf{\mu})^\top\mathbf{\Sigma}^{-1}({\bf{v}}_i-\mathbf{\mu})},\label{P}
\end{align}
where $\mathbf{\mu}$ is the mean vector and $\mathbf{\Sigma}$ is the corresponding covariance matrix. In this way, we only need to estimate $\mathbf{\mu}$ and $\mathbf{\Sigma}$ using the multivariate normal distribution, which reduces the computational burden of estimating irrelevant parameters. After this transformation, the missing data imputation for MTS becomes that filling in the missing sample $\mathbf{V}^m$ given the observed $\mathbf{V}^o$.

\subsection{Latent Gaussian Representation}
We then employ a mapping function $\mathcal{F}(\cdot)$ to transform the variables of ${\mathbf{v}}_i$ into a latent Gaussian representation with Gaussian copula, which is formulated as: 
\begin{equation}
    z_{ij} = \mathcal{F}_j(v_{ij}),
\end{equation}
where $z_{ij}$ is the latent variable corresponding to $v_{ij}$. $ \mathcal{F}_j(\cdot)$ represents the mapping function defined on the $j$-th variable in $\bf{V}$. Each $v_{ij} \in \bf{V}$ follows a multivariate Gaussian distribution $\mathcal{N}({\bf{\mu}}, {\bf{\Sigma}})$.

Based on the latent Gaussian representation ${\bf{z}}_i$, the prior probability distribution $\mathcal{P}({\bf{v}}_i|\mathbf{\mu},\mathbf{\Sigma})$ can be flexible represented by avoiding $\bf{\mu}$, which is formulated as: 
\begin{align}
    \mathcal{P}({{\bf{z}}_i|\mathbf{\Sigma}})=(2\pi)^{-\frac{M}{2}}\lvert{\Sigma}\rvert^{-\frac{1}{2}}e^{-\frac{1}{2} {\bf{z}}_i^\top\Sigma^{-1}{\bf{z}}_i},\label{p}
\end{align} 
where ${\bf{z}}_{i}=[z_{i1},z_{i2},...,z_{iM}]\in \mathbb{R}^M$ which can be obtained by the mapping ${\bf{v}}_i$. The mapping function $\mathcal{F}_j(\cdot)$ is implemented by: 
\begin{align}
    \mathcal{F}_{j}=\Phi^{-1} \circ P_{j}\label{F},
\end{align}
where $\Phi$ is the standard normal cumulative distribution function (CDF). $P_{j}$ is the CDF corresponding to the $j$-th variable.

The $\Phi$ is strictly monotonically increasing, making $\Phi^{-1}(x)$ exists, which is typically defined as the following:
\begin{align}
    \Phi(x)=\frac{1}{\sqrt{2\pi}}\int_{-\infty}^xe^{-\frac{t^2}{2}}dt.\label{standardCDF}
\end{align}

The $p_j$ manifests variations attributable to the diverse distributions of ${v}_{ij}$. For example, if ${v}_{ij}$ follows a normal distribution $N(\mu,\sigma)$, it makes the $u_{ij}=P_j({\bf{v}}_{ij})$ follows a uniform distribution on $[0,1]$, which is formulated as the following:
\begin{align}
    P_j(x)=\frac{1}{\sigma\sqrt{2\pi}}\int_{-\infty}^xe^{-\frac{(t-\mu)^2}{2\sigma^2}}dt.\label{PCDF}
\end{align}

Similarly, through $z_{ij}=\Phi^{-1}(u_{ij})$, the uniform distribution can be transformed into a standard normal distribution $N(0,1)$.

In order to map ${{\bf{z}}_{ij}}$ to ${{\bf{v}}_{ij}}$ back after imputing, we need define $\mathcal{F}_j^{-1}$ for all features. To accommodate the presence of outliers, ${v_{ij}}$ can take any value within the interval $[-\infty,+\infty]$. Therefore, for continuous variables, the probability density function is always greater than 0, and CDF is strictly monotonically increasing. By definition, the inverse of a strictly monotonically increasing function exists, hence $P_j^{-1}$ is present.

However, if the variable $v_{ij}\in\{1,2,...,K\}$, then CDF $P(v_{ij}\le k+\gamma)=P(v_{ij}\le k) ,k\in\{1,2,...K\},\gamma \in(0,1)$, meaning it only satisfies monotonically increasing. Therefore, we need to define a specific $P_j^{-1}$ to ensure the existence of the inverse mapping.

When considering a variable $v_{ij}$ that belongs to the set $\{1,2,...K\}$, where its CDF is $P_j$, and $\Phi(z_{ij})$ falls within the interval $[P_j(k-1),P_j(k)]$ for $k \in \{1,2,...,K\}$, we defines $\mathcal{F}_j^{-1}(z_{ij}) = P_j^{-1}(\Phi(z_{ij})) = k$.

So far, we define the hidden Gaussian vector mapping $\mathcal{F}_j(\cdot)$ and its inverse mapping $\mathcal{F}^{-1}_j(\cdot)$ for MTS data.

\subsection{Covariance Matrix Construction}
To fully explore the relation of latent variables, we leverage the maximum likelihood estimation (MLE) to estimate the correlation matrix $\bf{\Sigma}$, which is formulated as the following: 
\begin{equation}
\begin{array}{l}
\ell ({\bf{\Sigma }};{\bf{Z}}) 
 = c - \frac{1}{2}\log (|{\bf{\Sigma }}|) - \frac{1}{2}{\mathop{\rm Tr}\nolimits} \left[ {{{\bf{\Sigma }}^{ - 1}}\frac{1}{N}\sum\limits_{i = 1}^N {{{\bf{z}}_i}} {{\left( {{{\bf{z}}_i}} \right)}^ \top }} \right],\label{losss}
\end{array}
\end{equation}
where $c$ is a constant, $N$ denotes the number of patients.

When $\mathbf{Z}$ represents complete data, by maximize Equation \ref{losss}, we can derive the expression for $\mathbf{\Sigma}$ as follows:
\begin{align}
    \hat{\mathbf{\Sigma}}=\frac{1}{N}\sum_{i=1}^N\mathbf{z}_i\mathbf{z}_i^{\top}.\label{sigma}
\end{align}

It is worth noting that ${\bf{Z}}$ contains missing values. We maximize the joint probability of between $\mathbf{\Sigma}$ and observed value ${\bf{Z}}^o$ in the $\bf{Z}$.

\begin{align}
    \ell_{o} ({\bf{\Sigma}};{\bf{Z}}^o)=\frac{1}{N}\sum_{i=1}^N\int_{z\in{\bf{z}}_i^{{{o}}}}\mathcal{P}(z;\mathbf{\Sigma}_{o,o})\ dz,\label{update}
\end{align}
where ${\bf{\Sigma}}_{o,o}$ are submatrices of ${\bf{\Sigma}}$, with rows and columns corresponding to $({\bf{z}}_i^{o}, {\bf{z}}_i^{o})$. 
The optimal $\hat{\Sigma}$ can be obtained by maximizing the joint probability.
\begin{align}
    \hat{\bf{\Sigma}}=\arg\max\limits_{\bf{\Sigma}}{\ell_o ({\bf{\Sigma}};{\bf{Z}}^o)}.\label{argmax}
\end{align}

\subsection{Missing Value Estimation}
The missing values ${\bf{z}}_i^{m}$ for the $i$-th patient can be obtained based on its observed values ${\bf{z}}_i^{o}$ and covariance matrix $\bf{\Sigma}$:
\begin{align}
    \mathbb{E}[{\bf{z}}_i^{m}|{\bf{z}}_i^{o},{\bf{\Sigma}}]={\bf{\Sigma}}_{m,o}{\bf{\Sigma}}^{-1}_{o,o}{\bf{z}}_i^{o},\label{27}
\end{align}
where ${\bf{\Sigma}}_{m,o}$ and ${\bf{\Sigma}}_{o,o}$ are submatrices of ${\bf{\Sigma}}$, with rows and columns corresponding to $({\bf{z}}_i^{m}, {\bf{z}}_i^{o})$ and $({\bf{z}}_i^{o}, {\bf{z}}_i^{o})$, respectively. We then utilize the inverse mapping $\mathcal{F}_j^{-1}(\cdot)$ to transform $\bf{Z}$ back into ${\bf{V}}$.

\subsection{EM Optimization}
The EM algorithm has shown a remarkable ability to handle data with substantial missing values \cite{zhao2020missing}. To make a more accurate estimation of $\mathbf{\Sigma}$, we employ the EM algorithm to iteratively update the covariance matrix. Firstly, we initialize the missing part of the latent variable matrix $\mathbf{Z}^m$ to zero and compute the corresponding covariance matrix $\mathbf{\Sigma}^{(0)}$. The general procedure of the EM algorithm is as follows:
\subsubsection{E-Step} Utilizing $\mathbf{\Sigma}^{(t)}$ and $\mathbf{z}_i^o$, we recalculate $\mathbf{z}_i^m$ through Equation \ref{27} for all patients.We assume that the filled values are $\mathbf{z}_i^I$.
\begin{align}
    \mathbf{z}_i^I=\mathbb{E}[\mathbf{z}_i^m|{\bf{z}}^{o}_i,{\bf{\Sigma}}^{(t)}].
\end{align}
\subsubsection{M-Step} We calculate the updated covariance matrix $\hat{\mathbf{\Sigma}}^{(t+1)}$ using $G({\bf{\Sigma}}^{(t)},{\bf{z}}_{i}^{o_i})$. The function $G({\bf{\Sigma}}^{(t)},{\bf{z}}_{i}^{o_i})$ is related to Equation \ref{sigma} and is expressed as follows:
\begin{align}
    G({\bf{z}}^{I}_i,{\bf{z}}_{i}^{o})=\frac{1}{N}\sum_{i=1}^N\mathbb{E}[{\bf{z}}_i{\bf{z}}_i^\top|{\bf{z}}^{o}_i,{\bf{z}}^{I}_i].
\end{align}
\subsubsection{Scale} Given that Gaussian copulas correspond to the correlation matrix, we apply constraints on the iterative process of ${\bf{\Sigma}}$ through $P_{\varepsilon}$. The relationship is established as follows: $P_{\varepsilon} = {\bf{D}}^{-\frac{1}{2}}{\bf{\Sigma}} {\bf{D}}^{-\frac{1}{2}}$, where ${\bf{D}}$ is the diagonal matrix derived from ${\bf{\Sigma}}$. This transformation ensures that the constraints are effectively incorporated into the iterative estimation of ${\bf{\Sigma}}$.

\section{Experiments}
\subsection{Datasets Acquisition}
The experiments are conducted on three public real-world healthcare datasets, including Medical Information Mart for Intensive Care (MIMIC) \cite{johnson2023mimic}, PhysioNet 2012 Mortality Prediction Challenge (PhysioNet2012) \cite{silva2012predicting} and PhysioNet in Cardiology Challenge 2019 (PhysioNet2019) \cite{reyna2020early}. The statistical results of the three datasets are shown in Table \ref{table:1}. 

\begin{table}[h!]
	\caption{General information of four datasets used in this work}
	\label{table:1}
	\begin{center}
       \scalebox{0.92}{
		\renewcommand\arraystretch{1.1}
		\resizebox{1.0\linewidth}{!}{
			\begin{tabular}{cccc}%
				\hline%
				&MIMIC&PhysioNet-2012&PhysioNet-2019 \\
				\hline%
				\#Total Samples&20,940&11,988&36,409 \\
				\#Features&21&36&34 \\
				\#Sequence Length&31&48&18 \\
				Original Missing Rate&68\%&80\%&79\% \\
				\hline%
		\end{tabular}}
       }
	\end{center}
\end{table}

\begin{table*}[]
\caption{The comparison of TGC with other methods. The best performance per dataset is highlighted in bold. The results in brackets represent the average standard deviation of three experiments.}
\label{table2}
\renewcommand\arraystretch{1.1}
\resizebox{1.0\linewidth}{!}{
\begin{tabular}{cccccccccc}
\hline
\multirow{2}{*}{Model} & \multicolumn{3}{c}{MIMIC}                                                & \multicolumn{3}{c}{PhysioNet2012}                                        & \multicolumn{3}{c}{PhysioNet2019}                                        \\ \cline{2-10} 
                       & MAE                    & MRE                    & RMSE                   & MAE                    & MRE                    & RMSE                   & MAE                    & MRE                    & RMSE                   \\ \hline
LOCF \cite{overall2009last}                   & 0.547 (0.003)          & 0.722 (0.004)          & 0.842 (0.003)          & 0.447 (0.001)          & 0.632 (0.003)          & 0.864 (0.043)          & 0.520 (0.001)          & 0.698 (0.002)          & 0.800 (0.002)          \\
ISVD (2001) \cite{troyanskaya2001missing}                   & 0.525 (0.002)          & 0.693 (0.002)          & 0.748 (0.003)          & 0.520 (0.002)           & 0.735 (0.002)          & 0.814 (0.031)          & 0.519 (0.002)          & 0.696 (0.001)          & 0.753 (0.007)          \\
SI (2010) \cite{mazumder2010spectral}                     & 0.592 (0.001)          & 0.781 (0.001)          & 0.808 (0.002)          & 0.508 (0.001)          & 0.719 (0.001)          & 0.799 (0.034)          & 0.603 (0.002)          & 0.808 (0.000)              & 0.827 (0.006)          \\
BRITS (2018) \cite{cao2018brits}                  & 0.436 (0.002)          & 0.575 (0.002)          & 0.691 (0.003)          & 0.356 (0.001)          & 0.504 (0.001)          & 0.689 (0.038)          & 0.435 (0.002)          & 0.583 (0.002)          & 0.679 (0.006)          \\
KNN (2019) \cite{pujianto2019k}                    & 0.683 (0.002)          & 0.901 (0.002)          & 0.951 (0.003)          & 0.639 (0.001)          & 0.903 (0.003)          & 1.008 (0.026)          & 0.743 (0.003)          & 0.996 (0.003)          & 1.037 (0.006)          \\
GPVAE (2020) \cite{fortuin2020gp}                  & 0.568 (0.004)          & 0.749 (0.004)          & 0.791 (0.004)          & 0.544 (0.015)          & 0.769 (0.020)          & 0.834 (0.032)          & 0.569 (0.004)          & 0.762 (0.005)          & 0.798 (0.008)          \\
FiLM (2022) \cite{zhou2022film}                   & 0.596 (0.002) & 0.786 (0.002) & 0.835 (0.004) & 0.533 (0.002) & 0.754 (0.001) & 0.848 (0.031) & 0.580 (0.002)  & 0.777 (0.001) & 0.822 (0.006) \\
SCINet (2022) \cite{liu2022scinet}                 & 0.530 (0.002)  & 0.699 (0.003) & 0.756 (0.004) & 0.465 (0.002) & 0.657 (0.003) & 0.770 (0.034)  & 0.510 (0.004)  & 0.683 (0.004) & 0.747 (0.008) \\
SAITS (2023) \cite{du2023saits}                  & 0.401 (0.004)         & 0.528 (0.005)         & 0.631 (0.005)          & 0.352 (0.002)         & 0.498 (0.003)         & 0.677 (0.042)          & 0.418 (0.002)         & 0.560 (0.002)          & 0.660 (0.006)          \\
Koopa (2024) \cite{liu2024koopa}                  & 0.527 (0.005) & 0.695 (0.006) & 0.745 (0.005) & 0.476 (0.007) & 0.674 (0.011) & 0.768 (0.035) & 0.519 (0.007) & 0.696 (0.009) & 0.757 (0.009) \\
FreTS (2024) \cite{yi2024frequency}                  & 0.521 (0.008) & 0.688 (0.011) & 0.756 (0.009) & 0.488 (0.007) & 0.690 (0.009)  & 0.814 (0.043) & 0.477 (0.008) & 0.640 (0.011)  & 0.718 (0.010)  \\
TGC(ours)                   & \textbf{0.377 (0.001)} & \textbf{0.497 (0.002)} & \textbf{0.610 (0.002)} & \textbf{0.309 (0.001)} & \textbf{0.437 (0.002)} & \textbf{0.639 (0.042)} & \textbf{0.389 (0.001)} & \textbf{0.521 (0.001)} & \textbf{0.638 (0.006)} \\ \hline
\end{tabular}}
\end{table*}

\begin{figure*}[htbp]
	\centerline{\includegraphics[width=1.0\linewidth]{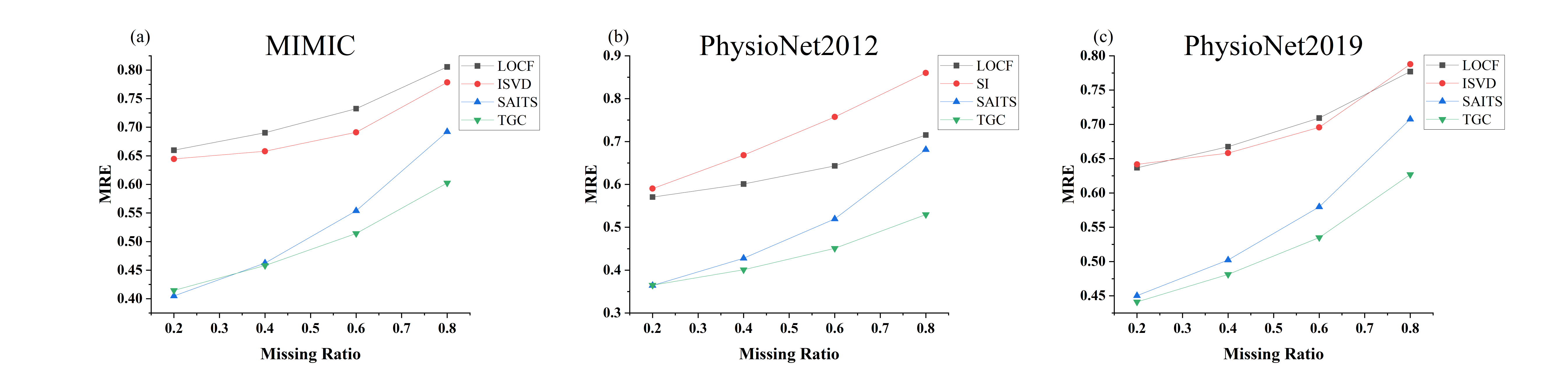}}
	\caption{The performance of representative models at various levels of data missingness. The missing ratio on the horizontal axis represents the proportion of observations removed from the test set randomly, not the total missing rate of the test set.}

	\label{fig2}
\end{figure*}

\subsection{Implementation Detail}
The data set is initially divided into the training set and the test set according to 80\% and 20\%. We randomly eliminate  $E\%$ of observed values in the test set and use these values as ground truth to evaluate the imputation performance of models (See \ref{advantageintro}). To fully leverage the computational power of statistical models designed for two-dimensional data, we reshaped the input data into a matrix format of size $N \times (TF)$. For deep learning-based models, we train the model on the training set and then perform interpolation on the masked test set to calculate the corresponding metrics. All the models are run through three rounds, with the average and variance of the corresponding metrics computed.  We have carefully calibrated the parameters of these models based on the original papers and made necessary adjustments to adapt them to the varying characteristics of our datasets. Following the prior work \cite{du2023saits, liu2024koopa},
MAE (mean absolute error), MRE (mean relative error), and RMSE (average square error of the radius) are used for evaluation.
We implement these models using the PyPOTS and fancyimpute packages and employ the Adam optimizer for optimization on an NVIDIA GeForce RTX 4060 GPU. 

\subsection{Comparison with Competing Methods}
\begin{table*}[]
	\caption{Ablation studies on the proposed model. The best performance is highlighted in bold. The results in brackets represent the average standard deviation of three experiments.}
 \label{table3}
	\renewcommand\arraystretch{1.1}
	\resizebox{1.0\linewidth}{!}{
	\begin{tabular}{cccccccccc}
		\hline
		\multirow{2}{*}{Model} & \multicolumn{3}{c}{MIMIC}                                                & \multicolumn{3}{c}{PhysioNet2012}                                        & \multicolumn{3}{c}{PhysioNet2019}                                        \\ \cline{2-10} 
		& MAE                    & MRE                    & RMSE                   & MAE                    & MRE                    & RMSE                   & MAE                    & MRE                    & RMSE                   \\ \hline
		TGC-U              & 0.760 (0.001)          & 1.002 (0.000)          & 1.003 (0.002)          & 0.709 (0.002)          & 1.003 (0.000)          & 1.01 (0.024)           & 0.746 (0.002)          & 1.000 (0.000)          & 0.995 (0.006)          \\
		U-ISVD                 & {0.529 (0.002)}          & {0.698 (0.003)}          &{0.752 (0.002)}          & {0.520 (0.002)}          & {0.735 (0.002)}          & {0.814 (0.031)}          & {0.519 (0.002)}          & {0.696 (0.001)}          & 0.753 (0.007)         \\
		TGC                    & \textbf{0.377 (0.001)} & \textbf{0.497 (0.002)} & \textbf{0.610 (0.002)} & \textbf{0.309 (0.001)} & \textbf{0.437 (0.002)} & \textbf{0.639 (0.042)} & \textbf{0.389 (0.001)} & \textbf{0.521 (0.001)} & \textbf{0.638 (0.006)} \\ \hline
	\end{tabular}}
\end{table*}

We compare the TGC with the existing SOTA imputation algorithms by reporting the mean performance of the tested models across multiple trials where 20\% to 80\% of the observation data are randomly removed from the test set, which are shown in Table \ref{table2}.
From Table \ref{table2}, we can see our TGC significantly outperforms other models, including statistical methods and deep learning-based methods on all databases.

TGC constructs a fully-connected covariance matrix to connect all the variables with all the time-points, thus it can learn complex dependencies and achieve the best results in MTS imputation.

\subsection{Imputation With Varing Missing Ratios}\label{advantageintro}
To demonstrate the model's effectiveness of handling the data with varying sampling densities, we further conduct experiments on the three datasets with different missing ratios, which is shown in Figure \ref{fig2}. In this experiment, all models were trained on datasets with the original missing rate (shown in Table \ref{table:1}), and the testing data was randomly removed by 20\% to 80\% to simulate different missing levels.
From the results, we observe that our TGC exhibits superior adaptability on the dataset with high missing ratios, indicating strong robustness. The statistical models perform poorly on large datasets with high missing rates, overall being inferior to deep learning models. Deep learning models are second only to TGC and higher than other models overall.

\subsection{Ablation Study}
In this subsection, to empirically evaluate the effectiveness of different modules of TGC, we have devised several variations of the TGC: (1) TGC-U, where the unfolding is replaced with a transformation of the form $(NT)\times F$, designed to differentiate it from the unfolding; (2) U-ISVD, which uses ISVD to fill in missing values after an unfolding operation. 

The comparative analysis of the various models is summarized in Table \ref{table3}. These findings reveal that the TGC-U model failed to converge, emphasizing the crucial role of unfolding in assisting the model to grasp the complex interdependencies among different variables across various time points. Additionally, the U-ISVD model exhibits a considerably greater loss than the TGC model, thereby reinforcing the superior performance of the Gaussian copula.

\section{Conclusion}
In this paper, we propose a TGC model for clinical MTS missing value estimation. We explore the cross-variable and temporal relationships based on the Gaussian latent representation. We employ an EM algorithm to iteratively refine the covariance matrix and impute missing values, thereby enhancing model adaptability. Extensive experiments validate the TGC model's superiority in clinical data imputation. Moreover, TGC exhibits remarkable robustness in handling test datasets with high missing rates.
\bibliographystyle{IEEEtran}
\bibliography{references}

\end{document}